\documentclass{article}

\PassOptionsToPackage{numbers, sort, compress}{natbib}

     \usepackage[preprint]{neurips_2024}

\usepackage[utf8]{inputenc} 
\usepackage[T1]{fontenc}    
\usepackage{hyperref}       
\usepackage{url}            
\usepackage{booktabs}       
\usepackage{amsfonts}       
\usepackage{nicefrac}       
\usepackage{microtype}      
\usepackage{xcolor}         

\usepackage{multirow}
\usepackage[ruled,vlined]{algorithm2e}
\usepackage{graphicx}
\usepackage{wrapfig}
\usepackage{amsmath,bm,bbm,upgreek}
\usepackage{amsthm}
\usepackage{amssymb}
\usepackage{pifont}
\usepackage{xcolor}
\usepackage{caption}
\usepackage{subcaption}
\usepackage{url}
\usepackage{comment, enumitem}

\newtheorem{lemma}{Lemma}[section]

\theoremstyle{definition}
\newtheorem{definition}{Definition}[section]

\newcommand{\rv}[1]{#1}

\title{A Gradient Analysis Framework for Rewarding Good and Penalizing Bad Examples in Language Models}

\author{%
       Yi-Lin Tuan\\
       University of California, Santa Barbara\\
       \texttt{ytuan@cs.ucsb.edu}
       \And
       William Yang Wang\\
       University of California, Santa Barbara\\
       \texttt{william@cs.ucsb.edu}
}

\begin{document}
    \maketitle
    
\begin{abstract}
    
Beyond maximum likelihood estimation (MLE), the standard objective of a language model (LM) that optimizes good examples probabilities, many studies have explored ways that also penalize bad examples for enhancing the quality of output distribution, including unlikelihood training, exponential maximizing average treatment effect (ExMATE), and direct preference optimization (DPO).
\rv{To systematically compare these methods and further provide a unified recipe for LM optimization, in this paper, we present a unique angle of gradient analysis of loss functions that {\it simultaneously reward good examples and penalize bad ones} in LMs.}
Through both mathematical results and experiments on CausalDialogue and Anthropic HH-RLHF datasets, we identify distinct functional characteristics among these methods.
We find that ExMATE serves as a superior surrogate for MLE, and that combining DPO with ExMATE instead of MLE further enhances both the statistical (5-7\%) and generative (+18\% win rate) performance.

\end{abstract}

\section{Introduction}

The optimization of language models (LM) has long relied on maximum likelihood estimation (MLE)~\cite{hochreiter1997long,cho2014learning,sutskever2014sequence}.
While MLE aims to concentrate probability distributions on {\it correct tokens} at each timestep, this approach has inherent limitations.
Solely optimizing for correct examples can lead to over-optimism on the referred token~\cite{jaques2020human} and unintended distribution (such as uniform) over unused tokens, regardless of the data scale.
Consequently, a paradigm shift has occurred, recognizing the need to consider both positive and negative examples in LM optimization.

To address the shortcomings of exclusively rewarding correct data, novel strategies have emerged, originating from binary classifiers~\cite{ren2003learning} and extending to sequential multi-class classifiers like LMs.
Techniques such as unlikelihood training~\cite{Welleck2020Neural} and exponential maximizing average treatment effect (ExMATE)~\cite{tuan-etal-2023-causaldialogue} introduce distinct loss functions and negative sample constructions to mitigate issues like repetition in text generation and enhance model response agility.
Meanwhile, generative adversarial networks (GANs) for LMs~\cite{ yu2017seqgan,tuan2019improving} and reinforcement learning from human feedback (RLHF)~\cite{ziegler2019fine,ouyang2022training} either directly takes machine generation as negative data or further annotates preference by humans to optimize the model via GAN or RL frameworks~\cite{goodfellow2014generative,ranzato2015sequence,schulman2017proximal,tuan2018proximal}.
Recently, direct preference optimization (DPO)~\cite{rafailov2024direct} streamlines the RLHF approach into a supervision loss objective, significantly reducing computational costs while maintaining efficacy.
These approaches collectively signify a broader shift towards optimizing LMs by simultaneously increasing the probability of preferred data and decreasing the probability of disliked data.

In this paper, we aim to systematically compare LM optimization methods that share the principle: {\it rewarding good and penalizing bad examples}.
Specifically, we address the following questions: (1) What are the essential differences among these methods in LM optimization? (2) \rv{Which method is more suitable for each scenario?} (3) Can we identify a superior optimization recipe based on \rv{mathematical} analysis?
To \rv{answer these questions}, we propose a gradient analysis approach tailored for frequently encountered LM scenarios, enabling us to mathematically estimate how each \rv{rewarding-good-penalizing-bad (RGPB)} method updates the LM output distribution and elucidate their distinct properties.
Additionally, we conduct experiments on datasets such as CausalDialogue~\cite{tuan-etal-2023-causaldialogue} and Anthropic HH-RLHF~\cite{bai2022training}, employing evaluations using statistical metrics and GPT4 assessments to verify our mathematical findings and validate the practical implications of our research.

\begin{figure}[t]
    \centering
    \includegraphics[width=1.\linewidth]{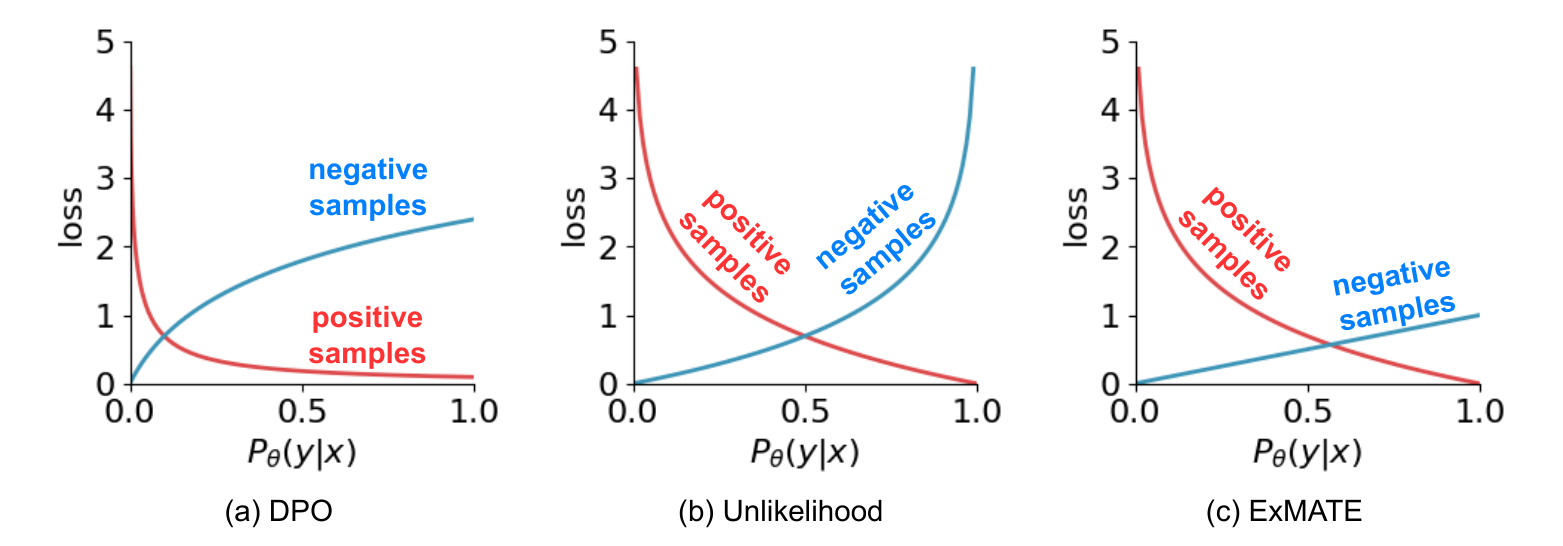}
    \caption{(a) DPO, (b) Unlikelihood, and (c) ExMATE loss functions when taking only either $P_\theta(y^+|x^+)$ (positive examples) or $P_\theta(y^-|x^-)$ (negative examples) as the control variables. We plot DPO in the case of $P_{ref}(\cdot)=1$, $\beta=1$, and $P_\theta(y^-|x^-)$ or $P_\theta(y^+|x^+)$ is 0.1 for easy visualization. Their function characteristics are different, thus making them suitable for difference use cases.}
    \label{fig:RGPB-functions}
\end{figure}

    \section{Related Work}

As rewarding good examples and penalizing bad ones has been \rv{one widely-used framework in LM research}, we discuss two \rv{major} differences in these works: (1) the method to construct negative data and (2) the method to optimize the LM using the negative data.
Moreover, we discuss (3) the difference of them from other lines of research, e.g., contrastive learning, that is often deemed similar.

{\bf Negative Data Construction.} Word2Vec~\cite{mikolov2013efficient,goldberg2014word2vec}, which aims to strengthen word embeddings, performs negative sampling by intentionally selecting incorrect positions for a word.
Unlikelihood training~\cite{Welleck2020Neural,li2020don}, which aims to prevent repetitions in text generation, uses already predicted words in the context as negative samples.
ExMATE~\cite{tuan-etal-2023-causaldialogue,tuan2024towards}, which aims to enhance LM response sensitivity to prior utterances or controls, constructs negative samples by replacing the context with a slightly incorrect predecessor.
GANs for LMs~\cite{li2017adversarial,yu2017seqgan,tuan2019improving} use the generator's outputs as negative samples.
RLHF~\cite{ziegler2019fine,jaques2020human,ouyang2022training,bai2022training} and its derivatives, such as DPO~\cite{rafailov2024direct}, IPO~\cite{azar2023general}, and KTO~\cite{ethayarajh2024kto}, collect human feedback to label pairwise preferred and rejected responses generated by a fine-tuned model, with the rejected responses serving as negative data.
In this work, we discuss often-seen cases in generative LMs, e.g., when the negative data that are also fluent language.
This requires the method to identify the nuanced difference between the positive and negative data.

{\bf Optimization Method.}
Word2Vec, GANs for LMs, and Unlikelihood training use a similar loss function long employed for optimizing binary classifiers~\cite{mikolov2013efficient,goodfellow2014generative,Welleck2020Neural}.
ExMATE~\cite{tuan-etal-2023-causaldialogue}, inspired by the average treatment effect and the directed acyclic graph structure of conversations~\cite{holland1986statistics,pearl2009causality}, proposes an exponential trick to linearize gradients to maintain language fluency.
DPO~\cite{rafailov2024direct} and its derivatives~\cite{azar2023general,ethayarajh2024kto,hong2024orpo}, also supervision loss functions, are firstly derived from RLHF~\cite{rafailov2024direct} to reduce the resource-intensive interactions in RL frameworks~\cite{ranzato2015sequence,bahdanau2016actor,kandasamy2017batch,tuan2018proximal}, relying on an assumed reward model~\cite{ouyang2022training,bai2022training} and Kullback–Leibler divergence regularization.
In this paper, we mainly discuss Unlikelihood, ExMATE, and DPO as they represent three distinct lines of research towards the same goal.
We discuss their function characteristics mathematically and empirically under the same setups.

{\bf Different from other Contrasts in ML.}
Contrastive learning~\citep{weinberger2009distance,gutmann2010noise,sohn2016improved,oord2018representation,chen2020simple} aims to learn similar representations for similar data points and vice versa.
The methods we discuss here, instead of learning representation space based on similarity among data points, aim to directly reshape the model output distribution based on each data point's intrinsic correctness, i.e., whether the input and output labels match.

    \section{Preliminary}
\label{sec:unification}

When training a generative LM $g$ with parameters set $\theta$, at each time step $t$, we feed the model an input sequence $x$ and a part of the expected output $y$.
The initial part of $y$, denoted as $y_{<t}$, indicating the first $t-1$ tokens in $y$.
The model predicts a probability distribution over a vocabulary set $\mathcal{V}$ per time step by:
\begin{equation}
    P_\theta(\cdot|x,y_{<t}) = softmax\bigl(g_\theta(x,y_{<t})\bigr)\,.
\end{equation}
We use $P_\theta$ in the rest to denote the LM, which is a combination of $g_\theta$ and the softmax function.

Assuming with training data $\mathcal{D}$ that involves correct text pairs $\{(x^+,y^+)_i\}_{i=1}^{|\mathcal{D}|}$, where the superscript $^+$ indicates the data sample is deemed correct, the model is often optimized by MLE as:
\begin{align}
    & \theta = \arg\min_\theta \mathcal{L}_{MLE}\,,\\
    & \mathcal{L}_{MLE} =  \underset{(x^+,y^+)\sim \mathcal{D}}{\mathbb{E}} \Biggl[ \frac{1}{T}\sum_{t=1}^T -\log P_\theta(y^+_t|x^+,y^+_{<t}) \Biggr]\,.
\end{align}
Minimizing $\mathcal{L}_{MLE}$ implies increasing the probability $P_\theta(y^+|x^+)$, as $P_\theta(y^+|x^+) = \prod_{t=1}^T P_\theta(y^+_t|x^+,y^+_{<t})$.
If assuming the model capacity and data scale are sufficiently large, the optimal can be achieved.

Nonetheless, without those strong assumptions and computation supports, studies have shown that considering negative examples $(x^-,y^-)$ can improve model performance~\cite{mikolov2013efficient,goldberg2014word2vec,Welleck2020Neural,li2020don,tuan-etal-2023-causaldialogue,rafailov2024direct,azar2023general,ethayarajh2024kto}.
We refer these methods as types of {\it Rewarding-Good-and-Penalizing-Bad} training loss and RGPB for short in the later sections.
In this paper, we discuss three types of RGPB methods: DPO~\cite{rafailov2024direct}, Unlikelihood training~\cite{Welleck2020Neural}, and ExMATE~\cite{tuan-etal-2023-causaldialogue}.

With our definitions of positive examples $(x^+,y^+)$ and negative examples $(x^-,y^-)$ from training data $\mathcal{D}$, the objectives of DPO, Unlikelihood (UL for brevity), and ExMATE are to update the model parameters $\theta$ to respectively minimize the loss functions $\mathcal{L}_{DPO}$, $\mathcal{L}_{UL}$, $\mathcal{L}_{ExMATE}$.
Our unification of negative sampling and GANs with Unlikelihood is in Appendix~\ref{apx:unification}.
Their formulations are:
\begin{align}
    \mathcal{L}_{DPO} & = -\mathbb{E}_\mathcal{D} \biggl[ \log \sigma \biggl( \beta \log \frac{P_\theta(y^+|x^+)}{P_{ref}(y^+|x^+)} - \beta \log \frac{P_\theta(y^-|x^-)}{P_{ref}(y^-|x^-)} \biggr) \biggr] \,,\\
    \mathcal{L}_{UL} & = -\mathbb{E}_\mathcal{D} \Biggl[ \frac{1}{T}\sum_{t=1}^T \log P_\theta(y^+_t|x^+,y^+_{<t}) + \beta \log \Bigl( 1-P_\theta(y^-_t|x^-,y^-_{<t})\Bigr)\Biggr]\,,\\
    \mathcal{L}_{ExMATE} & = -\mathbb{E}_\mathcal{D} \Biggl[ \frac{1}{T}\sum_{t=1}^T\log P_\theta(y^+_t|x^+,y^+_{<t}) - \beta \exp \biggl( \frac{1}{T}\sum_{t=1}^T\log P_\theta(y^-_t|x^-,y^-_{<t}) \biggr)\Biggr]\,.
\end{align}

For gradient analysis in the next sections, we first derive their gradient with respect to the model parameters $\theta$ (All the proofs and detailed derivations of this paper are in Appendix~\ref{apx:proofs}-\ref{apx:derive-grad-sameyt}).
The gradient of a loss function $\mathcal{L}$ is then used to update the model $\theta\leftarrow \theta - \nabla_\theta \mathcal{L}$.
We present the gradients here with notations $f_\theta^+ := P_\theta(y^+|x^+)$, $f_\theta^- := P_\theta(y^-|x^-)$, and $f_{ref}:=P_{ref}$ for brevity:
\begin{align}
    \nabla_\theta \mathcal{L}_{DPO} & = - \beta \mathbb{E}_\mathcal{D} \Bigl[ \sigma \Bigl(\beta \log \frac{f_\theta^-}{f_{ref}^-} - \beta \log \frac{f_\theta^+}{f_{ref}^+}\Bigr) \Bigl( \frac{\nabla_\theta f_\theta^+}{f_\theta^+} - \frac{\nabla_\theta f_\theta^-}{f_\theta^-} \Bigr) \Bigr]\,,\label{eq:grad-dpo}\\
    \nabla_\theta \mathcal{L}_{UL} & =  - \mathbb{E}_\mathcal{D} \Bigl( \frac{\nabla_\theta f_\theta^+}{f_\theta^+} + \beta \frac{-\nabla_\theta f_\theta^-}{1-f_\theta^-} \Bigr)\,,\label{eq:grad-ul}\\
    \nabla_\theta \mathcal{L}_{ExMATE} & = - \mathbb{E}_\mathcal{D} \Bigl( \frac{\nabla_\theta f_\theta^+}{f_\theta^+} -\beta\nabla_\theta f_\theta^- \Bigr)\,.\label{eq:grad-exmate}
\end{align}
\vspace{-10pt}
    \section{Factors Impact RGPB Gradients in Generative Language Models}

\subsection{Language Model Properties}
\label{subsec:grad-lm-properties}
Before diving into the gradient analysis, we ask what are the properties of generative LMs and what makes their gradients different from the usual classification problem.

\vspace{-5pt}
\paragraph{Multiple Time Steps.} We are fundamentally tackling every time steps instead of the whole $P(y^+|x^+)$ and $P(y^-|x^-)$.
We highlight the goal of an RGPB method for language models: We feed the model with different inputs $x^+$ and $x^-$, and ask the model to respectively optimize the probability of the token $y^+_t$ and deoptimize the probability of the token $y^-_t$ for every time steps $t$.

\vspace{-5pt}
\paragraph{Multiple Classes.}
Generating responses from a language model is a sequence of multi-class classification problems, i.e., the model predicts a probability distribution $P(\cdot|x,y_{<t})\in[0,1]^{|\mathcal{V}|}$ at each time step $t$ over the whole vocabulary set $\mathcal{V}$.
The generation result is often based on the whole probability distribution (e.g., Softmax sampling, nucleus sampling), not just a single token probability.
Therefore, beyond $y^+_t$ and $y^-_t$, other tokens in $\mathcal{V}$ can have impact.

\vspace{-5pt}
\paragraph{Literal Similarity.}
\rv{Being} natural language, \rv{$y^+$ and $y^-$ may use the same tokens} at some time steps.
For example, \rv{when $y^+$ and $y^-$ are respectively ``\underline{I'm doing great} today'' and ``\underline{I'm doing great} yesterday'', they are mostly the same with minor word changes; when they are respectively ``We enjoy in \underline{hiking}'' and ``They love \underline{hiking}'', they use single same word.}
Whether $y^+$ and $y^-$ share some same tokens plays a vital role in the gradient.

\subsection{Information and Gradient Differences between positive and negative samples}

Besides the characteristics of language models in Section~\ref{subsec:grad-lm-properties}, as shown in Equations~\ref{eq:grad-dpo}-\ref{eq:grad-exmate} in Section~\ref{sec:unification}, $f_\theta^+$, $f_\theta^-$, $\nabla_\theta f_\theta^+$, $\nabla_\theta f_\theta^-$ are the keys to determine the gradient for model update.

Among them, the information difference and gradient difference can have high impact.
We define them as following:

\begin{definition}\label{def:info-s}
    \it (Information Difference) $|\epsilon| := |f_\theta^+ - f_\theta^-|$. The difference between data samples $(x^+,y^+)$ and $(x^-,y^-)$ in terms of their probability masses for any $\theta$.
\end{definition}

\begin{definition}\label{def:grad-s}
    \it (Gradient Difference) $\|\nabla_\theta f_\theta^+ - \nabla_\theta f_\theta^-\|_p$, where $p$ indicates p-norm.
\end{definition}

\begin{lemma}
    \it In LMs with softmax function for final prediction, the Gradient Difference is determined by (1) the softmax distribution difference $\|P_\theta(\cdot|x^+,y^+_{<t}) - P_\theta(\cdot|x^-,y^-_{<t})\|_p$ (we use it as the gradient difference in the rest of the paper) and (2) the sameness of target output tokens. Proved in Appendix~\ref{apx:proofs}.
\end{lemma}

Information difference and gradient difference have a similar form, but gradient difference considers the probability distribution over the whole vocabulary set instead of single token probability mass.
This is also the reason that gradient difference for each time step $t$ is considered separately and information difference is the aggregation of all time steps probability masses.

These two variables and the above language model properties are the keys for gradient analysis in the next section.

\section{RGPB Gradient Analysis in Generative Language Models}
\label{sec:grad-analysis}

With the {\bf Multiple Time Steps} and {\bf Literal Similarity} properties of LMs, we split the gradient analysis into two parts: (1) gradient at time step $t$ that $y^+_t \neq y^-_t$, and (2) gradient at time step $t$ that $y^+_t = y^-_t$.

Furthermore, we drop the negation sign of Equations~\ref{eq:grad-dpo}-\ref{eq:grad-exmate} to consider the case of gradient ascent, and denote that (1) $P^+(\cdot) := P_\theta(\cdot|x^+,y^+_{<t})$, $P^-(\cdot) := P_\theta(\cdot|x^-,y^-_{<t})$, and (2) $f_\theta^+ = u\in[0,1]$, $f_\theta^- = u+\epsilon\in[0,1]$ for brevity, where $|\epsilon|$ is the defined information difference and it is an important factor for gradients.

\subsection{For time steps $t$ that $y^+_t \neq y^-_t$.}
\label{sec:grad-diffyt}

With chain rule, the gradients can be split into the two parts: (1) From the loss function to the logits (i.e., $\frac{\partial \mathcal{L}}{\partial g_\theta}$), and (2) from the logits to the model parameters $\theta$ (i.e., $\frac{\partial g_\theta}{\partial \theta}$).
We assume here that the gradient difference is small, i.e., $P^+_t \approx P^-_t \in [0,1]^{|\mathcal{V}|}$, so their logits' derivatives are approximately the same and denoted as $\nabla_\theta \zeta \in \mathbb{R}^{|\mathcal{V}|\times|\theta|} :=\frac{\partial g_\theta^+}{\partial \theta} \approx \frac{\partial g_\theta^-}{\partial \theta}$.

We rewrite Equations~\ref{eq:grad-dpo}-\ref{eq:grad-exmate} as followings and first look into the gradients that flow through a token $z\in \mathcal{V}$ when $z=y_t^+$ or $z=y_t^-$.
\begin{align}\small
    \nabla_\theta \mathcal{L}_{DPO} & = \nabla_\theta \zeta \cdot \frac{\beta(u+\epsilon)^{\beta}}{(u+\epsilon)^\beta+u^\beta} \left\{\begin{array}{l} 1-P^+(y^+_t)+P^-(y^+_t)\text{, if }z = y^+_t\\-P^+(y^-_t)-(1-P^-(y^-_t))\text{, if }z = y^-_t\\ \end{array}\right .\label{eq:grad_dpo_ytdiff_yt}\\
    \nabla_\theta \mathcal{L}_{UL} & = \nabla_\theta \zeta \cdot \left\{\begin{array}{l} 1-P^+(y^+_t)+\beta \frac{P^-(y^-_t)P^-(y^+_t)}{1-P^-(y^-_t)}\text{, if }z = y^+_t\\-P^+(y^-_t)-\beta P^-(y^-_t)\text{, if }z = y^-_t\\ \end{array}\right .\label{eq:grad_ul_ytdiff_yt}\\
    \nabla_\theta \mathcal{L}_{ExMATE} & = \nabla_\theta \zeta \cdot \left\{\begin{array}{l} 1-P^+(y^+_t)+\beta P^-(y^-_t)P^-(y^+_t)\text{, if }z = y^+_t\\-P^+(y^-_t)-\beta P^-(y^-_t)(1-P^-(y^-_t))\text{, if }z = y^-_t\\ \end{array}\right .\label{eq:grad_exmate_ytdiff_yt}
\end{align}
From Equations~\ref{eq:grad_dpo_ytdiff_yt}-\ref{eq:grad_exmate_ytdiff_yt}, all methods result in non-negative gradients for $z=y^+_t$ and non-positive gradients for $z=y^-_t$, indicating that whenever $y^+_t \neq y^-_t$, the model outputs are updated {\bf as expectation to raise the probability of $y^+_t$ and lower the probability of $y^-_t$}.
However, their updating rates $|\nabla_\theta|$ and stop criterion are different:
(1) DPO's $|\nabla_\theta|$ increases with $\epsilon$ but is always not infinity and becomes zero when $\epsilon\rightarrow -u$ ($f_\theta^-\rightarrow 0$).
(2) Unlikelihood's $|\nabla_\theta|$ increases with $P^-(y_t^-)$ (often correlated with $\epsilon$), but $|\nabla_\theta|$ for $y^+_t$ explodes.
Moreover, $|\nabla_\theta|$ for $y^+_t$ only becomes zero when $P^+(y^+_t)=1$.
(3) ExMATE's $|\nabla_\theta|$ for both $y^+_t$ and $y^-_t$ increase with $P^-(y_t^-)$ and are bounded. The $|\nabla_\theta|$ for $y^+_t$ also only becomes zero when $P^+(y^+_t)=1$.

The gradients for tokens $z\in \mathcal{V}$ except for $y_t^+$ and $y_t^-$:
\begin{align}
    \nabla_\theta \mathcal{L}_{DPO} & = \nabla_\theta \zeta \cdot \frac{\beta (u+\epsilon)^{\beta}}{(u+\epsilon)^\beta+u^\beta} \Bigl(-P^+(z)+P^-(z)\Bigr)\approx 0\,,\label{eq:grad_dpo_ytdiff_z}\\
    \nabla_\theta \mathcal{L}_{UL} & = \nabla_\theta \zeta \cdot \Bigl(-P^+(z)+ \beta \frac{P^-(y^-_t)}{1-P^-(y^-_t)}P^-(z)\Bigr)\,,\label{eq:grad_ul_ytdiff_z}\\
    \nabla_\theta \mathcal{L}_{ExMATE} & = \nabla_\theta \zeta \cdot \Bigl(-P^+(z)+\beta P^-(y^-_t)P^-(z)\Bigr)\,.\label{eq:grad_exmate_ytdiff_z}
\end{align}
From Equations~\ref{eq:grad_dpo_ytdiff_z}-\ref{eq:grad_exmate_ytdiff_z} and with the small gradient difference assumption that $P^+_t\approx P^-_t$, (1) DPO does not update probability of non-referred tokens (neither $y^+_t$ nor $y^-_t$), always only compensating $P^-(y_t^-)$ for $P^+(y_t^+)$.
(2) Unlikelihood stops the gradient when $P^-(y^-_t)=\frac{1}{1+\beta}$. When $P^-(y^-_t)>\frac{1}{1+\beta}$, Unlikelihood reduces $P^-(y^-_t)$ to increase the probabilities of non-referred tokens and $P^+(y^+_t)$; when $P^-(y^-_t)<\frac{1}{1+\beta}$, Unlikelihood also compensates probabilities of non-referred tokens to raise $P^+(y^+_t)$.
(3) ExMATE only decays $P^-(y^-_t)$ to compensate for $P^+(y^+_t)$ when $P^-(y^-_t)\rightarrow \frac{1}{\beta}$. When $P^-(y^-_t)\rightarrow 0$, ExMATE compromises $P(z)$ for $P(y^+_t)$.

Above all, DPO aims to only exchange probabilities of $y^+_t$ and $y^-_t$ and stops to increase $y^+_t$ when $y^-_t$ reaches zero probability.
On the other hand, ExMATE prioritizes to increase the probability of $y^+_t$ and only stops when $y^+_t$ reaches the highest probability by compensating both $y^-_t$ and all other tokens $z$.
Unlikelihood also aims to both increase the probability of $y^+_t$ until it reaches the highest probability and always decay the probability of $y^-_t$, but it also always compensate the probability of all other tokens $z$ for either $y^+_t$ or $y^-_t$.

\begin{figure}[t]
    \centering
    \includegraphics[width=\linewidth]{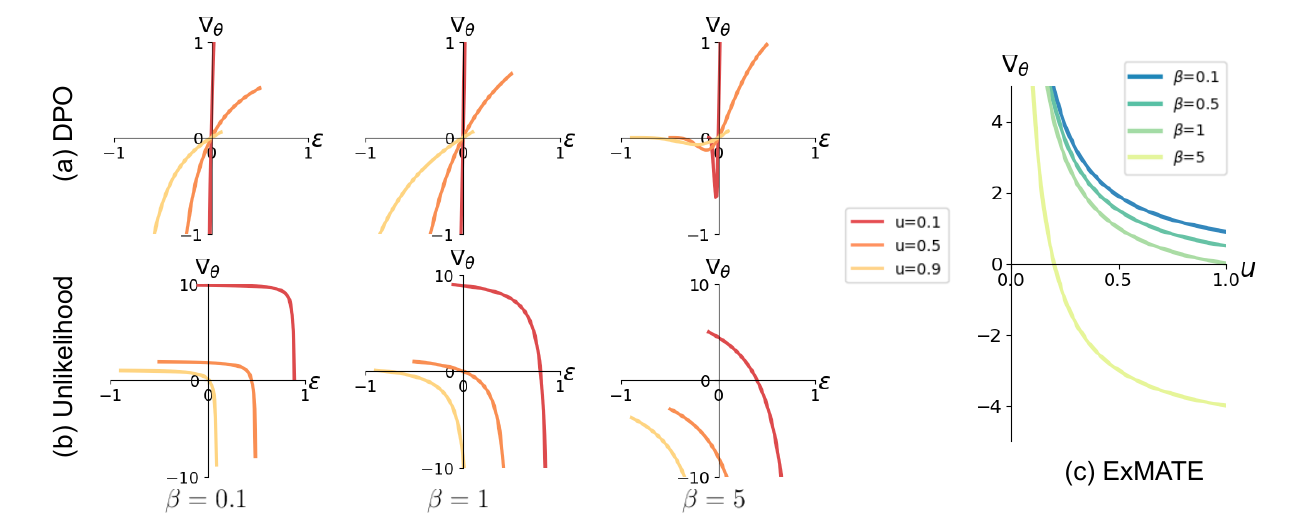}
    \caption{The estimated gradients of DPO, Unlikelihood, and ExMATE for time steps $t$ that $y^+_t = y^-_t$.}
    \label{fig:same_token_grads}
\end{figure}

\subsection{For time steps $t$ that $y^+_t = y^-_t$.}
\label{sec:grad-sameyt}

Another cases in LMs is when $y^+_t = y^-_t$.
Since now $y^+_t = y^-_t:=y_t$ and we assume that $P^+_t\approx P^-_t$, we can approximate that $\nabla_\theta f_\theta^+ \approx \nabla_\theta f_\theta^- =: \nabla_\theta f$.

The gradients become the following and we can interpret that when the gradient is positive, both $P^+(y^+_t)$ and $P^-(y^-_t)$ will raise, and vice versa.
\begin{align}
    \nabla_\theta \mathcal{L}_{DPO} & = \nabla_\theta f \cdot \frac{\beta (u+\epsilon)^{\beta-1}\epsilon}{((u+\epsilon)^\beta+u^\beta)u} \,,\label{eq:grad-dpo-st}\\
    \nabla_\theta \mathcal{L}_{UL} & = \nabla_\theta f \cdot \frac{1-(1+\beta)u-\epsilon}{u(1-u-\epsilon)}\,,\label{eq:grad-ul-st}\\
    \nabla_\theta \mathcal{L}_{ExMATE} & = \nabla_\theta f \cdot \Bigl( \frac{1}{u}-\beta\Bigr)\,.\label{eq:grad-exmate-st}
\end{align}
(1) DPO's gradients (Figure~\ref{fig:same_token_grads}(a) and Equation~\ref{eq:grad-dpo-st}) highly depend on $\epsilon$.
DPO increases $P^+(y_t)$ and $P^-(y_t)$ when $\epsilon>0$ ($f_\theta^- > f_\theta^+$) and decreases them when $\epsilon<0$.
This leads to the model {\bf decaying both $f_\theta^+$ and $f_\theta^-$ when reaching $\epsilon<0$}, which may not be desired in every cases.
Moreover, when $f_\theta^+\approx f_\theta^-$ (or $\epsilon\rightarrow 0$), the model does not learn things.
(2) Unlikelihood (Figure~\ref{fig:same_token_grads}(b) and Equation~\ref{eq:grad-ul-st}) decays $P^+(y_t)$ and $P^-(y_t)$ when $\epsilon > 1-(1+\beta)u$ and the decay rate explodes as $\epsilon\rightarrow 1-u$.
Meanwhile, when $u$ is lower, Unlikelihood mostly increases $P^+(y_t)$ and $P^-(y_t)$; when $u$ is higher, it mostly reduces $P^+(y_t)$ and $P^-(y_t)$.
This high rate of negative gradients is a reason for easily broken language after training.
(3) Differently, ExMATE's gradients (Figure~\ref{fig:same_token_grads}(c) and Equation~\ref{eq:grad-exmate-st}) only depend on $u$ and are always positive when $u<1/\beta$.
The positive gradients also have higher values than the negative ones.
Therefore, ExMATE mostly prioritizes to increase $P^+(y_t)$ and $P^-(y_t)$.

\subsection{Summary}
Overall, DPO mathematically (1) does not optimize $P(y^+|x^+)$ if $P(y^-|x^-)$ is already minimized and (2) tends to decrease all probabilities, so it is suitable for model optimization when some probability decays are acceptable, $P(y^+|x^+)$ is not required to be optimized, and $\epsilon$ is not nearly zero.
Unlikelihood mathematically aims to optimize both $P(y^+|x^+)$ and $P(y^-|x^-)$ by updating the probability of other tokens and the gradients are often large or exploded to facilitate the update, so it is more suitable for cases when minimizing $P(y^-|x^-)$ is nearly important as maximizing $P(y^+|x^+)$ and the literal similarity of $y^+$ and $y^-$ is lower.
ExMATE aims to optimize $P(y^+|x^+)$ by first reducing $P(y^-|x^-)$ and then reducing the probability of other tokens if $P(y^-|x^-)$ is already minimized. Moreover, its gradients are mostly bounded and less depend on $\epsilon$.
It is preferred when the $\epsilon\rightarrow 0$ or when maximizing $P(y^+|x^+)$ should be prioritized.

    \section{Experiments}

Beyond mathematical results, we are interested in RGPB methods' empirical results on real data and off-the-shelf LMs.
We first verify whether our assumptions in gradient analysis of information and gradient differences hold in real scenarios, e.g., diverse perfection levels of models (pre-trained or randomly initialized) and distinct relationships between the positive and negative samples.
We then ask: Can any of the RGPB methods generalize to different cases? What are their empirical properties? \rv{Do} they match the \rv{mathematical} results?

We will first describe our settings and then present the results.

\paragraph{Tasks.}
We experimented on two text generation datasets with different relationships between the positive and negative examples:
(1) {\bf CausalDialogue}~\cite{tuan-etal-2023-causaldialogue}, a conversation dataset with multiple $(x^+,y^+)$ and $(x^-,y^-)$ pairs extracted from the utterance directed acyclic graphs (DAG).
The $y^+$ and $y^-$ are the same while the $x^+$ and $x^-$ have only subtle difference.
The goal is to maximize $P_\theta(y^+|x^+)$ while minimizing $P_\theta(y^-|x^-)$.
This task is expected to have small information difference ($\epsilon\rightarrow 0$).
(2) {\bf Anthropic HH-RLHF}~\cite{bai2022training}, a dataset of human-machine dialogues ended with paired human preferred response and human rejected response.
This task is expected to have higher information difference between the positive and negative examples.
Also, since both $y^+$ and $y^-$ are machine generation instead of human written responses, we expect that a lower $P_\theta(y^+|x^+)$ is acceptable.

\paragraph{Methods.}
We compare DPO, Unlikelihood, and ExMATE with their coefficient $\beta$ tuned among $\{0.05,0.1,0.5,1,5\}$.
We also train models using MLE as a reference of LM performance without considering negative examples.
The MLE fine-tuned model is also referred to as SFT in the following to match the naming conventions of RLHF literatures~\cite{ouyang2022training}.
For the initial models and training recipe, we follow prior works~\cite{tuan-etal-2023-causaldialogue,rafailov2024direct}.
We fine-tune T5 models~\cite{raffel2020exploring} on CausalDialogue for five epochs, fix learning rate as 1e-5, allow a maximum of 128 input tokens and put no restriction on the output length.
We use Pythia-2.8B and Pythia-6.9B~\cite{biderman2023pythia} on Anthropic HH-RLHF for one epoch with fix learning rate 5e-7.
Our implementations follow their open-source codebases: \url{https://github.com/Pascalson/CausalDialogue} and \url{https://github.com/eric-mitchell/direct-preference-optimization}.

\paragraph{Evaluation.}
We primarily evaluate a model by perplexity and agility~\cite{tuan-etal-2023-causaldialogue, hong2024orpo}.
Perplexity, defined as $\exp[-\frac{1}{T}\sum_{t=1}^{T}\log P_\theta(y^+_t|x^+,y^+_{<t})]$, is to quantify the certainty of a model for $(x^+,y^+)$ and is used to automatically estimate a model's fluency.
Agility, defined as $f^+_\theta - f^-_\theta$ (which is also $-\epsilon$), is to quantify whether the model successfully rewarding-good while penalizing bad examples in their probability masses.
In addition to statistical evaluation, we evaluate by GPT4 the quality of sampled responses from the trained models.

\subsection{The values of information and gradient differences in real scenarios}

\begin{figure}[t]
    \centering
    \includegraphics[width=\linewidth]{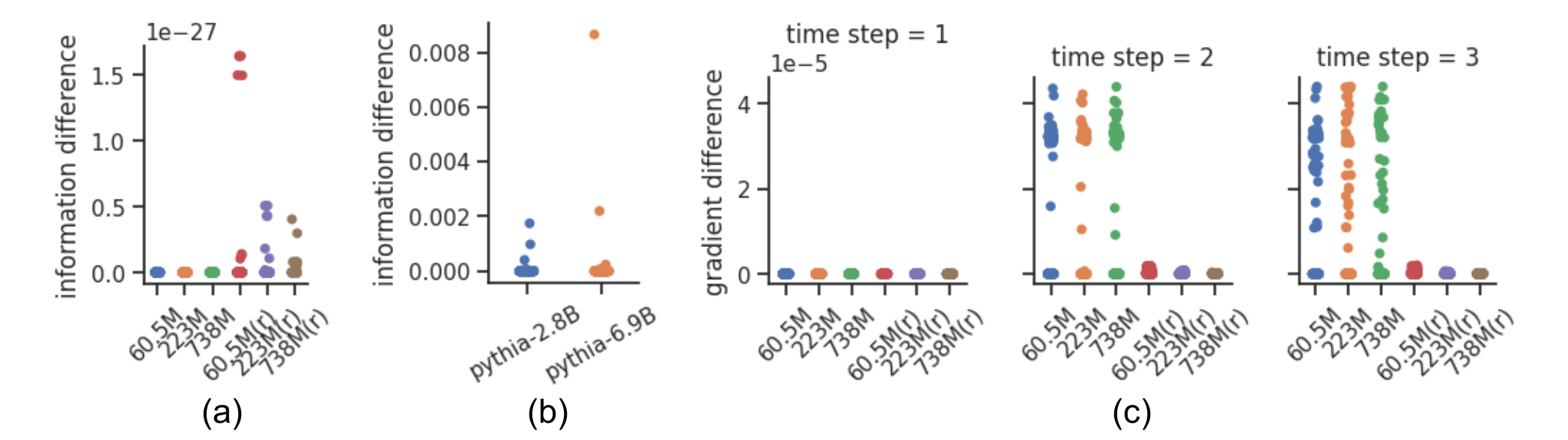}
    \caption{{\bf (a)} All model's information differences on CausalDialogue are nearly zero (<1e-26). {\bf (b)} information differences on Anthropic HH-RLHF are higher than on CausalDialogue. {\bf (c)} All model's gradient differences on CausalDialogue and the first three time steps. All are small, especially for the first time step, randomly initialized models, and larger number of parameters.}
    \label{fig:all-similarities}
\end{figure}

We first empirically verify whether our assumptions in Section~\ref{sec:grad-analysis} of the information difference and gradient difference (Definitions~\ref{def:info-s} and \ref{def:grad-s}) between $(x^+,y^+)$ and $(x^-,y^-)$ hold: The gradient difference is mostly low and negligible and the information difference can be nearly zero or higher.
We test 8 situations in total, including CausalDialogue with pretrained and randomized T5-small (60.5M), T5-base (223M), T5-large (738M) models, Anthropic Helpful and Harmless Dialogue with pretrained Pythia-2.8B and Pythia-6.9B.

Results are shown in Figure~\ref{fig:all-similarities}(a)(b), where each point is the value for a pair of $(x^+,y^+)$ and $(x^-,y^-)$.
On CausalDialogue, the information difference is nearly zero for all model sizes, even though slightly higher when using non-pretrained models.
Differently, Anthropic HH-RLHF with large LMs has higher information difference.
The key of information difference is the literal similarity between $(x^+,y^+)$ and $(x^-,y^-)$.

The gradient difference, as in Figure~\ref{fig:all-similarities}(c), is low for every generation steps, especially the first step on CausalDialogue, and it is always zero for Anthropic HH-RLHF, since the $x^+$ and $x^-$ are always the same.
The reason of the increasing gradient difference along time steps is that dialogue responses often have similar openings and the literal difference will accumulate along generation steps.

\begin{figure}[t]
    \centering
    \includegraphics[width=\linewidth]{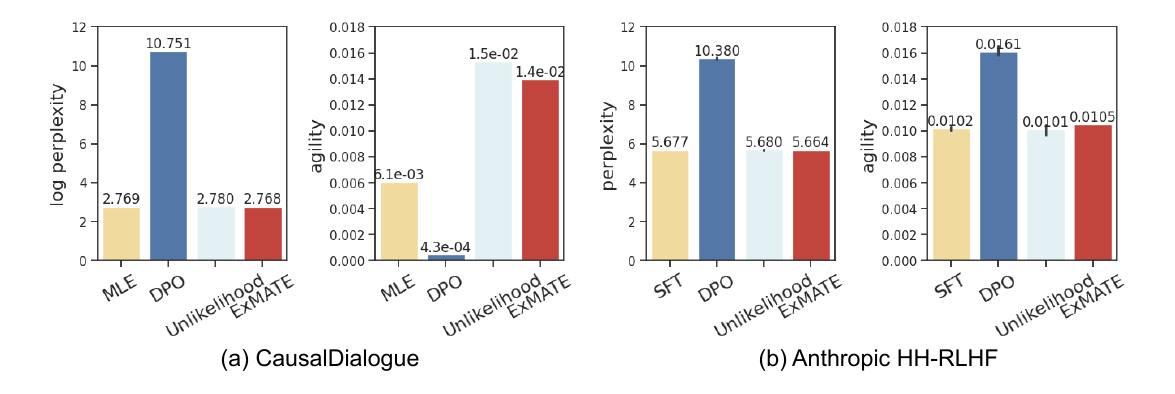}
    \caption{
    {\bf (a)} Perplexity (log scale) and agility of MLE, DPO, Unlikelihood, and ExMATE on CausalDialogue. Unlikelihood improves agility, DPO degrades both, and ExMATE is preferred for improving both.
    {\bf (b)} Perplexity and agility of SFT(MLE), DPO, Unlikelihood, and ExMATE on Anthropic HH-RLHF. DPO achieves high agility by compromising perplexity; ExMATE improves both metrics by small values.}
    \label{fig:main-results}
\end{figure}

\subsection{Comparing RGPB methods in the case of low information difference.}
\label{subsec:results-high-similarity}

Since CausalDialogue has low information and gradient differences, Figure~\ref{fig:main-results}(a) shows empirical results in a real scenarios discussed in Section~\ref{sec:grad-sameyt}.
DPO introduces almost zero gradients and results in high perplexity and zero agility, giving no convergence and effective learning.
On the other hand, since Unlikelihood often introduces gradients to decay probabilities, the perplexity is higher than simply using MLE.
However, the good news is, as the probabilities are overall small, Unlikelihood does not introduce unwanted exploded gradient is this case.
ExMATE simultaneously improve perplexity and introduces the second highest agility score, reflecting the fact in gradient analysis that it prioritize to increase probability of $(x^+,y^+)$.

\subsection{Comparing RGPB methods in the case of higher information difference.}

\begin{wrapfigure}[14]{r}{0.5\textwidth}
  \vspace{-20pt}
  \begin{center}
    \includegraphics[width=0.5\textwidth]{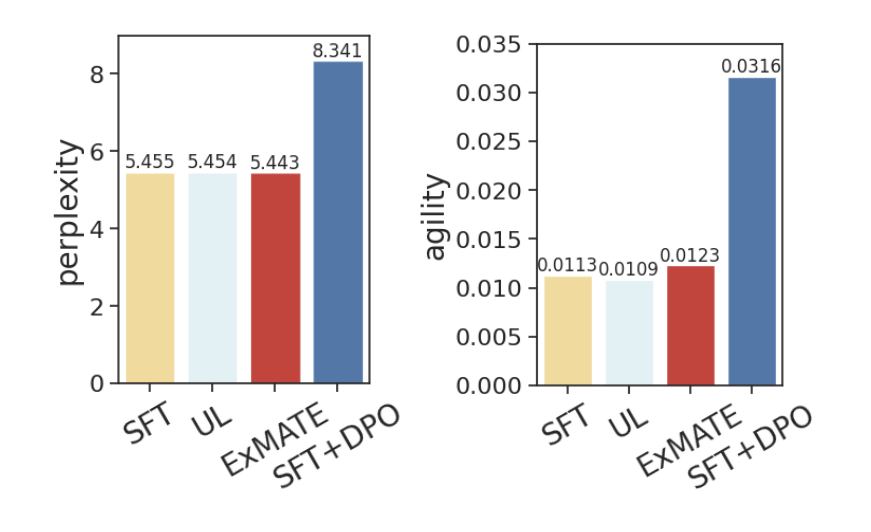}
  \end{center}
  \vspace{-10pt}
  \caption{Fine-tuned Pythia 6.9B by SFT, Unlikelihood (UL), ExMATE, or SFT+DPO.}
  \label{fig:ahh-7B-results}
\end{wrapfigure}

Another case we mainly discuss is Anthropic HH-RLHF that shows higher information differences and matches the case of gradient analysis in Section~\ref{sec:grad-diffyt}.
The results in Figure~\ref{fig:main-results}(b) shows that DPO achieves the highest agility and compromises much perplexity.
This is expected that DPO can perform better in this case compared to situations with lower information differences due to no zero gradient issue.
However, DPO still tends to decay the probabilities as our gradient analaysis.
On the other hand, ExMATE acheives the second best agility (but only slightly higher agility compared to other methods) and the lowest perplexity.

We also test with larger model Pythia 6.9B and followed prior work to improve DPO by first fine-tuning the model with SFT (called SFT+DPO) and plot the results in Figure~\ref{fig:ahh-7B-results}.
The results that both perplexity and agility of all methods are improved, showing that these RGPB methods are all scalable to model size and SFT+DPO still retains the property of DPO that compromising perplexity.

\subsection{Discussion of New Methods: ExMATE with SFT or ExMATE with DPO?}

To find a better recipe of RGPB beyond MLE, we first observe that (1) MLE/SFT, Unlikelihood, and ExMATE have many similar trends while ExMATE consistently achieves lowest perplexity and higher agility, and (2) DPO's trend is an outlier and, as prior work discussed, may require the model to be first fine-tuned by SFT to reach certain performance.

\begin{wrapfigure}[16]{r}{0.5\textwidth}
  \vspace{-12pt}
  \begin{center}
    \includegraphics[width=0.5\textwidth]{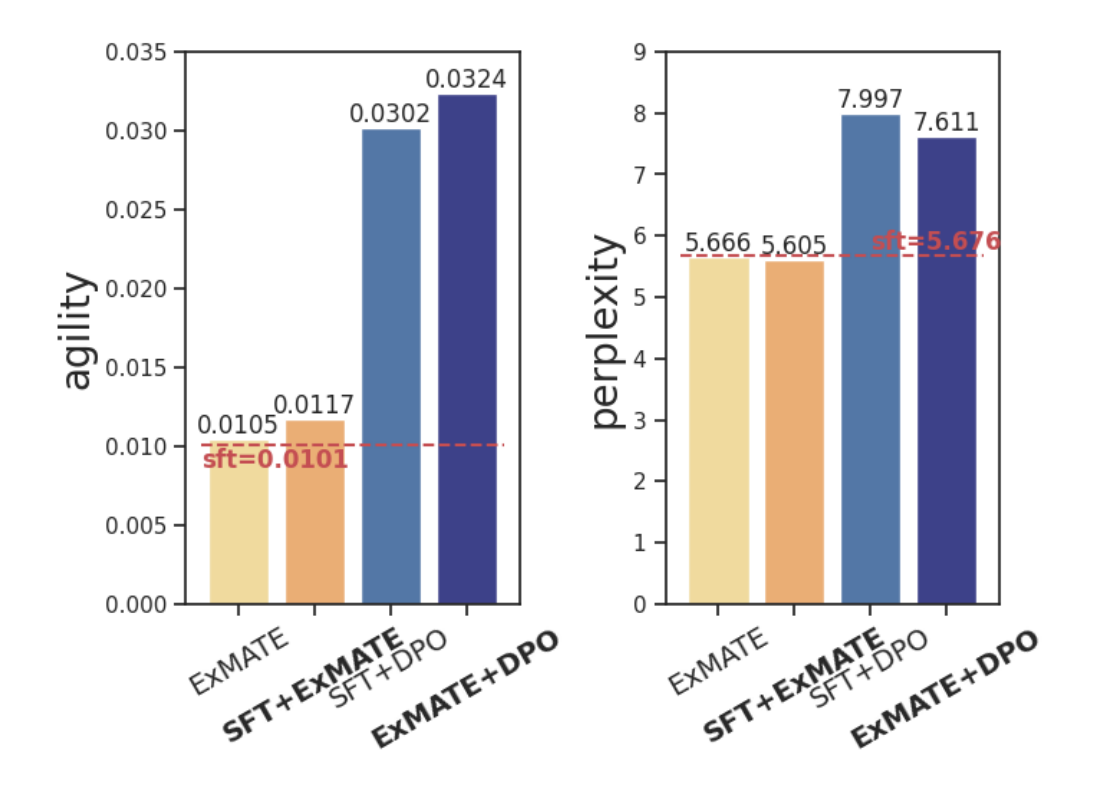}
  \end{center}
  \vspace{-10pt}
  \caption{ExMATE vs. SFT+ExMATE and SFT+DPO vs. ExMATE+DPO.}
  \label{fig:ahh-newmethods}
\end{wrapfigure}

Therefore, we compare: (1) ExMATE vs SFT+ExMATE by replacing the DPO in SFT+DPO framework with ExMATE since ExMATE is an overall best performed RGPB method, and (2) SFT+DPO vs ExMATE+DPO by replacing the SFT stage with ExMATE since ExMATE shares similar perplexity and agility trends with SFT but is better.
The results are shown in Figure~\ref{fig:ahh-newmethods} and the red line is the SFT result for reference.
The plots clearly show that:
(1) All methods improve agility.
(2) ExMATE+SFT improves both agility and perplexity while DPO+SFT sacrifice much perplexity for agility, showing the different properties of ExMATE and DPO.
(3) ExMATE+DPO improves both agility and perplexity compared to SFT+DPO, indicating ExMATE can also be a better surrogate for MLE.
The results also demonstrate that developing a better supervised loss function can simultaneously have preferred statistical properties (high agility and low perplexity) and can be empirically aggregated with other methods for performance boost.

\begin{wraptable}[9]{r}{0.48\linewidth}
  \vspace{-10pt}
  \caption{GPT4 evaluation results.}\label{tab:gpt4eval}
  \vspace{-15pt}
  \begin{tabular}{lcc}\\\toprule
     & win & lose \\\midrule
    ExMATE vs DPO & 0.47 & 0.40 \\  \midrule
    ExMATE vs SFT+DPO & 0.32 & 0.36 \\  \midrule
    ExMATE+DPO vs SFT+DPO & 0.56 & 0.38 \\
    \bottomrule
  \end{tabular}
\end{wraptable}

\subsection{Validating evaluation results by GPT4 and human judgements}

\rv{This paper focuses on analyzing the statistical effect of RGPB methods, which can be shown by perplexity and agility metrics.
They are the direct objectives of RGPB methods and provide us an overview of the learned output distribution properties.
In addition, we also evaluate the generated responses by GPT4 and human judgements to gain other types of understanding.}
\rv{Such judgements can also give us an understanding of the link between GPT4 or human evaluation with the statistical properties reflected in perplexity and agility.}
Specifically, we give GPT4 a conversation and two generated responses and ask GPT4 to choose the better response.
The responses are generated from \rv{Pythia-2.8B models trained on HH-RLHF data} using (1) ExMATE vs DPO, (2) ExMATE vs SFT+DPO, and (3) ExMATE+DPO vs SFT+DPO, of which statistical results are shown in Figure~\ref{fig:main-results}(b).
\rv{Table~\ref{tab:gpt4eval} presents the GPT4 evaluation results.}
\rv{To verify the trustfulness of the GPT4 evaluation restuls, we ask} human annotators the same task and gain 0.842 Cohen's kappa, indicating strong agreement between GPT4 and human ratings~\cite{mchugh2012interrater}.
From the results, we interpret that agility and perplexity are both important metrics:
(1) When only one of them is better, their sampling results may be indifference. Worse perplexity often leads to less fluent response and better agility often leads to less helpful or safe responses (ExMATE vs DPO and ExMATE vs SFT+DPO).
(2) When both are better, the sampling results also show improvements (ExMATE+DPO vs SFT+DPO).
    \section{Discussion}
In this paper, we gather and analyze loss functions that essentially share the same goal: simultaneously rewarding good data samples and penalizing bad data samples in LM output distribution.
The representative methods we discuss are: unlikelihood training, exponential maximizing ATE (ExMATE), and direct preference optimization (DPO) as their proposed time order.
We provide a novel perspective to consider the characteristics of generative LMs into gradient analysis: the multiple time steps, multiple classes, and literal similarity.
Our approach splits gradient analysis into two primary cases: (1) When $y^+_t\neq y^-_t$ and (2) $y^+_t=y^-_t$.
{\bf Generalization Performance.} From both mathematical results and experiments, we conclude that DPO although can significantly increase agility, which is defined as the gap between probability masses of positive and negative samples, it largely compromises perplexity and fails to introduce effective gradients when the initial information difference is small.
In contrast, ExMATE consistently enlarges agility and prevents the probability drops in the same time across situations, but with a relatively small improvements.
These demonstrate that agility and perplexity are not necessarily trade-offs but the quantity of improvements needs to be enhanced.
{\bf A Better Recipe.} With further experiments taking ExMATE as a surrogate of SFT or DPO, we suggest a more unified optimization method is to first train LMs by ExMATE instead of MLE.
If the test case does not require sufficiently high probability of given positive examples, then we use DPO to do further fine-tuning.

{\bf Limitations.} We approach the difference of RGPB methods from the view of gradient analysis in common LM scenarios, so the results may not suitable for all possible cases, such as $(x^-,y^-)$ is broken language without any similarity with $(x^+,y^+)$ or single step binary classifier.
Also, we focus on whether the LM output distribution is as desired: having higher probability of the positive examples and lower probability of the negative examples, which are reflected in perplexity and agility metrics. However, other metrics of generation quality, such as language diversity, informativeness, factuality, which relate less to the output distribution are not specifically discussed.
They are deemed orthogonal to the function characteristics and can be included as the difference between positive and negative examples.

{\bf Broader Impacts and Future Work.} As this work discusses, RGPB methods aim to optimize generative LMs instead of MLE.
Therefore, while applying the suggested method can train higher quality of generative models, it can be purposely used to generate harmful responses.
We expect the proposed gradient analysis approach can facilitate future improvement of RGPB methods on both perplexity and agility, e.g., first designing the desired gradients and then transferring back to the loss function.
We expect that future work can be inspired to rethink the importance of perplexity and agility in different tasks and choose the more suitable RGPB method.
Furthermore, as recent advancements in other aspects of LMs, such as model architectures and system frameworks, we hope more advancements in loss functions for LM can address how to learn a more desired output distrition beyond MLE.

    \bibliographystyle{plainnat}\small
    \bibliography{main}

    \newpage
    \appendix
    \section{Methods Unification}\label{apx:unification}

\paragraph{Negative Sampling.}
As an early approach for language modeling, specifically for word2vec model~\cite{mikolov2013efficient}, negative sampling is derived from the goal~\cite{goldberg2014word2vec}:
\begin{equation}
    \arg\max_\theta \prod_{(x,y)\in \mathcal{D}} P(D=1|x,y; \theta) \prod_{(x,y)\in\mathcal{D}'} P(D=0|x,y;\theta)
\end{equation}
Using our unified notations, the equation becomes:
\begin{equation}
    \arg\max_\theta \log P_\theta(y^+|x^+) + \log(1-P_\theta(y^-|x^-))
\end{equation}

\paragraph{Generative Adversarial Networks.}
The discriminator of a GAN~\cite{goodfellow2014generative,yu2017seqgan} is optimized to identify its input data is real or fake by:
\begin{equation}\label{eq:GAN}
    \arg\max_D \mathbb{E}_{\mathcal{D}_{real}} \log D(y^+|x^+) + \mathbb{E}_{\mathcal{D}_{fake}} \log[1-D(y^-|x^-)]
\end{equation}
The discriminator is normally appended with a sigmoid function, so $D(\cdot)$ outputs a scalar within $[0,1]$.
It is therefore possible to view $D(\cdot)$ as a probability function with parameters set $\theta$.
The resulting formula is replacing the $D$ in Equation~\ref{eq:GAN} with $P_\theta$, thus almost the same as unlikelihood training~\cite{Welleck2020Neural,li2020don}.
The three differences are: (1) the negative samples are generated by the generator in GAN and rule-based constructed in Unlikelihood training, (2) the second term in Unlikelihood training is multiplied with a coefficient to adjust the negative samples weight during training, (3) Unlikelihood training consider multiple time steps and multiple classes.

    \section{Proof of Lemma 4.1}
\label{apx:proofs}
In LMs with softmax function before the final prediction, the output probability distribution can be written as $\frac{e^{z_j}}{\sum_{k=1}^K e^{z_k}}$, where the $z_1$,$z_2$,...$z_K$ are the predicted logit for each token in the vocabulary set $\mathcal{V}$ with size $K$.
The gradient with respect to $\theta$ from a logit $z_i$ is:
\begin{align}
    & \nabla_\theta f_\theta = \sum_{i=1}^K (\frac{\partial}{\partial z_i} \frac{e^{z_j}}{\sum_{k=1}^K e^{z_k}}) \frac{\partial z_i}{\partial \theta}\,,\\
    & \frac{\partial}{\partial z_i} \frac{e^{z_j}}{\sum_{k=1}^K e^{z_k}} = \left\{\begin{array}{l} softmax_\theta^j(1 - softmax_\theta^j)\text{, if }i=j\\-softmax_\theta^j softmax_\theta^i\text{, if }i\neq j \end{array}\right .
\end{align}
Since the gradient of a given sample ($j$-th token here) is a function of $softmax_\theta$ and $j$, the difference of gradients between two given samples is determined by (1) their softmax distribution difference and (2) whether the target tokens (j) are the same.
    \section{Gradients Derivations of Section~\ref{sec:unification}}
\label{apx:grad}

\subsection{Derivation of $\nabla_\theta \mathcal{L}_{DPO}$}

\begin{equation}
    \begin{split}
        & \nabla_\theta \mathcal{L}_{DPO} = \nabla_\theta -\mathbb{E}_\mathcal{D} \biggl[ \log \sigma \biggl( \beta \log \frac{P_\theta(y^+|x^+)}{P_{ref}(y^+|x^+)} - \beta \log \frac{P_\theta(y^-|x^-)}{P_{ref}(y^-|x^-)} \biggr) \biggr]\\
        & = -\mathbb{E}_\mathcal{D} \biggl[ \sigma \biggl( - \beta \log \frac{P_\theta(y^+|x^+)}{P_{ref}(y^+|x^+)} + \beta \log \frac{P_\theta(y^-|x^-)}{P_{ref}(y^-|x^-)} \biggr) \biggr] \nabla_\theta \Bigl(\beta \log P_\theta(y^+|x^+) - \beta \log P_\theta(y^-|x^-) \Bigr)\\
        & = -\mathbb{E}_\mathcal{D} \biggl[ \sigma \biggl( - \beta \log \frac{P_\theta(y^+|x^+)}{P_{ref}(y^+|x^+)} + \beta \log \frac{P_\theta(y^-|x^-)}{P_{ref}(y^-|x^-)} \biggr) \biggr] \beta \Bigl( \nabla_\theta \log P_\theta(y^+|x^+) - \nabla_\theta \log P_\theta(y^-|x^-) \Bigr)\\
        & = -\mathbb{E}_\mathcal{D} \biggl[ \sigma \biggl( - \beta \log \frac{P_\theta(y^+|x^+)}{P_{ref}(y^+|x^+)} + \beta \log \frac{P_\theta(y^-|x^-)}{P_{ref}(y^-|x^-)} \biggr) \biggr] \beta \Bigl( \frac{\nabla_\theta P_\theta(y^+|x^+)}{P_\theta(y^+|x^+)} - \frac{\nabla_\theta P_\theta(y^-|x^-)}{P_\theta(y^-|x^-)} \Bigr)
    \end{split}
\end{equation}

\subsection{Derivation of $\nabla_\theta \mathcal{L}_{UL}$}

\begin{equation}
    \begin{split}
        & \nabla_\theta \mathcal{L}_{UL} = \nabla_\theta -\mathbb{E}_\mathcal{D} \Biggl[ \frac{1}{T}\sum_{t=1}^T \log P_\theta(y^+_t|x^+,y^+_{<t}) + \beta \log \Bigl( 1-P_\theta(y^-_t|x^-,y^-_{<t})\Bigr)\Biggr]\\
        & \approx \nabla_\theta -\mathbb{E}_\mathcal{D}  \frac{1}{T} \Biggl[ \log P_\theta(y^+|x^+) + \beta \log \Bigl( 1-P_\theta(y^-|x^-)\Bigr)\Biggr]\\
        & = -\mathbb{E}_\mathcal{D}  \frac{1}{T} \Biggl[ \frac{\nabla_\theta P_\theta(y^+|x^+)}{P_\theta(y^+|x^+)} + \beta \frac{-\nabla_\theta P_\theta(y^-|x^-)}{1-P_\theta(y^-|x^-)} \Biggr]
    \end{split}
\end{equation}

\subsection{Derivation of $\nabla_\theta \mathcal{L}_{ExMATE}$}

\begin{equation}
    \begin{split}
        & \nabla_\theta \mathcal{L}_{ExMATE} = \nabla_\theta -\mathbb{E}_\mathcal{D} \Biggl[ \frac{1}{T}\sum_{t=1}^T\log P_\theta(y^+_t|x^+,y^+_{<t}) - \beta \exp \biggl( \frac{1}{T}\sum_{t=1}^T\log P_\theta(y^-_t|x^-,y^-_{<t}) \biggr)\Biggr]\\
        & = \nabla_\theta -\mathbb{E}_\mathcal{D} \Biggl[  \frac{1}{T} \log P_\theta(y^+|x^+) - \beta \exp \biggl( \frac{1}{T} \log P_\theta(y^-|x^-) \biggr)\Biggr]\\
        & = -\mathbb{E}_\mathcal{D} \Biggl[  \frac{1}{T} \frac{\nabla_\theta P_\theta(y^+|x^+)}{P_\theta(y^+|x^+)} - \beta \exp\biggl(\frac{1}{T}\biggr) \nabla_\theta P_\theta(y^-|x^-)\Biggr]\\
        & = -\frac{1}{T} \mathbb{E}_\mathcal{D} \Biggl[ \frac{\nabla_\theta P_\theta(y^+|x^+)}{P_\theta(y^+|x^+)} - \beta' \nabla_\theta P_\theta(y^-|x^-)\Biggr]
    \end{split}
\end{equation}

\newpage
\section{Derivations of Section~\ref{sec:grad-diffyt} when $y^+_t\neq y^-_t$}
\label{apx:derive-grad-diffyt}

With an assumption that $P_{ref}(\cdot)=1$ and denote that (1) $P^+(\cdot) := P_\theta(\cdot|x^+,y^+_{<t})$, $P^-(\cdot) := P_\theta(\cdot|x^-,y^-_{<t})$, (2) $f_\theta^+ = u\in[0,1]$, $f_\theta^- = u+\epsilon\in[0,1]$, and (3) $\nabla_\theta \zeta \in \mathbb{R}^{|\mathcal{V}|\times|\theta|} :=\frac{\partial g_\theta^+}{\partial \theta} \approx \frac{\partial g_\theta^-}{\partial \theta}$ for brevity.

\begin{equation}
    \begin{split}
        \nabla_\theta \mathcal{L}_{DPO} & = -\mathbb{E}_\mathcal{D} \biggl[ \sigma \biggl( - \beta \log \frac{P_\theta(y^+|x^+)}{P_{ref}(y^+|x^+)} + \beta \log \frac{P_\theta(y^-|x^-)}{P_{ref}(y^-|x^-)} \biggr) \biggr] \beta \Bigl( \frac{\nabla_\theta P_\theta(y^+|x^+)}{P_\theta(y^+|x^+)} - \frac{\nabla_\theta P_\theta(y^-|x^-)}{P_\theta(y^-|x^-)} \Bigr)\\
        & = -\mathbb{E}_\mathcal{D} \biggl[ \sigma \biggl( \beta \log(u+\epsilon) - \beta \log u  \biggr) \biggr] \beta \sum_{t=1}^T \Bigl( \frac{\nabla_\theta P^+(y^+_t)}{P^+(y^+_t)} - \frac{\nabla_\theta P^-(y^-_t)}{P^-(y^-_t)} \Bigr)\\
        & = -\mathbb{E}_\mathcal{D} \sigma \biggl( \log(\frac{u+\epsilon}{u}^\beta)\biggr) \beta \sum_{t=1}^T \Bigl( \frac{\nabla_\theta P^+(y^+_t)}{P^+(y^+_t)} - \frac{\nabla_\theta P^-(y^-_t)}{P^-(y^-_t)} \Bigr)\\
        & = -\mathbb{E}_\mathcal{D} \frac{(u+\epsilon)^\beta}{(u+\epsilon)^\beta+u^\beta} \beta \sum_{t=1}^T \nabla_\theta \zeta \cdot \left\{\begin{array}{l} \frac{-P^+(y^+_t)P^+(z)}{P^+(y^+_t)}-\frac{-P^-(y^-_t)P^-(z)}{P^-(y^-_t)}\text{, if }z\neq y^+_t, z\neq y^-_t\\\frac{P^+(y^+_t)(1-P^+(y^+_t))}{P^+(y^+_t)}-\frac{-P^-(y^-_t)P^-(y^+_t)}{P^-(y^-_t)}\text{, if }z = y^+_t\\\frac{-P^+(y^+_t)P^+(y^-_t)}{P^+(y^+_t)}-\frac{P^-(y^-_t)(1-P^-(y^-_t))}{P^-(y^-_t)}\text{, if }z = y^-_t\\ \end{array}\right .\\
        & = -\mathbb{E}_\mathcal{D} \frac{(u+\epsilon)^\beta}{(u+\epsilon)^\beta+u^\beta} \beta \sum_{t=1}^T \nabla_\theta \zeta \cdot \left\{\begin{array}{l} -P^+(z)+P^-(z)\text{, if }z\neq y^+_t, z\neq y^-_t\\(1-P^+(y^+_t))+P^-(y^+_t)\text{, if }z = y^+_t\\-P^+(y^-_t)-(1-P^-(y^-_t))\text{, if }z = y^-_t\\ \end{array}\right .\\
    \end{split}
\end{equation}

\begin{equation}
    \begin{split}
        \nabla_\theta \mathcal{L}_{UL} & = -\mathbb{E}_\mathcal{D}  \frac{1}{T} \sum_{t=1}^T \Biggl[ \frac{\nabla_\theta P^+(y^+_t)}{P^+(y^+_t)} + \beta \frac{-\nabla_\theta P^-(y^-_t)}{1-P^-(y^-_t)} \Biggr]\\
        & = -\mathbb{E}_\mathcal{D}  \frac{1}{T} \sum_{t=1}^T \nabla_\theta \zeta \cdot \left\{\begin{array}{l} \frac{-P^+(y^+_t)P^+(z)}{P^+(y^+_t)}-\beta \frac{-P^-(y^-_t)P^-(z)}{1-P^-(y^-_t)}\text{, if }z\neq y^+_t, z\neq y^-_t\\\frac{P^+(y^+_t)(1-P^+(y^+_t))}{P^+(y^+_t)}-\beta \frac{-P^-(y^-_t)P^-(y^+_t)}{1-P^-(y^-_t)}\text{, if }z = y^+_t\\\frac{-P^+(y^+_t)P^+(y^-_t)}{P^+(y^+_t)}-\beta \frac{P^-(y^-_t)(1-P^-(y^-_t))}{1-P^-(y^-_t)}\text{, if }z = y^-_t\\ \end{array}\right .\\
        & = -\mathbb{E}_\mathcal{D}  \frac{1}{T} \sum_{t=1}^T \nabla_\theta \zeta \cdot \left\{\begin{array}{l} -P^+(z)+\beta \frac{P^-(y^-_t)P^-(z)}{1-P^-(y^-_t)}\text{, if }z\neq y^+_t, z\neq y^-_t\\(1-P^+(y^+_t))+\beta \frac{P^-(y^-_t)P^-(y^+_t)}{1-P^-(y^-_t)}\text{, if }z = y^+_t\\-P^+(y^-_t)-\beta P^-(y^-_t)\text{, if }z = y^-_t\\ \end{array}\right .\\
    \end{split}
\end{equation}

\begin{equation}
    \begin{split}
        \nabla_\theta \mathcal{L}_{ExMATE} & = \nabla_\theta -\mathbb{E}_\mathcal{D} \Biggl[ \frac{1}{T}\sum_{t=1}^T\log P_\theta(y^+_t|x^+,y^+_{<t}) - \beta \exp \biggl( \frac{1}{T}\sum_{t=1}^T\log P_\theta(y^-_t|x^-,y^-_{<t}) \biggr)\Biggr]\\
        & \approx -\frac{1}{T} \sum_{t=1}^T \mathbb{E}_\mathcal{D} \Biggl[ \frac{\nabla_\theta P^+(y^+_t)}{P^+(y^+_t)} - \beta \nabla_\theta P^-(y^-_t)\Biggr]\\
        & = \frac{1}{T} \sum_{t=1}^T \nabla_\theta \zeta \cdot \left\{\begin{array}{l} \frac{-P^+(y^+_t)P^+(z)}{P^+(y^+_t)}+\beta P^-(y^-_t)P^-(z)\text{, if }z\neq y^+_t, z\neq y^-_t\\\frac{P^+(y^+_t)(1-P^+(y^+_t))}{P^+(y^+_t)}+\beta P^-(y^-_t)P^-(y^+_t)\text{, if }z = y^+_t\\\frac{-P^+(y^+_t)P^+(y^-_t)}{P^+(y^+_t)}-\beta P^-(y^-_t)(1-P^-(y^-_t))\text{, if }z = y^-_t\\\end{array}\right .\\
        & = \frac{1}{T} \sum_{t=1}^T \nabla_\theta \zeta \cdot \left\{\begin{array}{l} -P^+(z)+\beta P^-(y^-_t)P^-(z)\text{, if }z\neq y^+_t, z\neq y^-_t\\(1-P^+(y^+_t))+\beta P^-(y^-_t)P^-(y^+_t)\text{, if }z = y^+_t\\-P^+(y^-_t)-\beta P^-(y^-_t)(1-P^-(y^-_t))\text{, if }z = y^-_t\\ \end{array}\right .
    \end{split}
\end{equation}

\newpage
\section{Derivations of Section~\ref{sec:grad-sameyt} when $y^+_t = y^-_t := y_t$}
\label{apx:derive-grad-sameyt}

We first prove that $\nabla_\theta f_\theta^+ \approx \nabla_\theta f_\theta^-$ given that $y^+_t = y^-_t$ and $P^+_t\approx P^-_t$:
\begin{equation}
    \begin{split}
        \nabla_\theta f_\theta^+ & = \left\{\begin{array}{l} -P^+(y_t)P^+(z)\text{, if }z\neq y_t\\P^+(y_t)(1-P^+(y_t))\text{, if }z = y_t\\ \end{array}\right .\\
        & \approx \left\{\begin{array}{l} -P^-(y_t)P^-(z)\text{, if }z\neq y_t\\P^-(y_t)(1-P^-(y_t))\text{, if }z = y_t\\ \end{array}\right .\\
        & = \nabla_\theta f_\theta^-\,.
    \end{split}
\end{equation}

We then reduce the gradients of DPO, Unlikelihood, and ExMATE as following:
\begin{equation}
    \begin{split}
        \nabla_\theta \mathcal{L}_{DPO} & = - \beta \mathbb{E}_\mathcal{D} \Bigl[ \sigma \Bigl(\beta \log \frac{f_\theta^-}{f_{ref}^-} - \beta \log \frac{f_\theta^+}{f_{ref}^+}\Bigr) \Bigl( \frac{\nabla_\theta f_\theta^+}{f_\theta^+} - \frac{\nabla_\theta f_\theta^-}{f_\theta^-} \Bigr) \Bigr]\\
        & = - \beta \mathbb{E}_\mathcal{D} \Bigl[ \frac{(u+\epsilon)^\beta}{(u+\epsilon)^\beta+u^\beta} \Bigl( \frac{\nabla_\theta f_\theta}{u} - \frac{\nabla_\theta f_\theta}{u+\epsilon} \Bigr) \Bigr]\\
        & = - \beta \mathbb{E}_\mathcal{D} \Bigl[ \frac{(u+\epsilon)^{\beta-1}\epsilon}{((u+\epsilon)^\beta+u^\beta)u} \nabla_\theta f_\theta \Bigr]\\
    \end{split}
\end{equation}

\begin{equation}
    \begin{split}
        \nabla_\theta \mathcal{L}_{UL} & =  - \mathbb{E}_\mathcal{D} \Bigl( \frac{\nabla_\theta f_\theta^+}{f_\theta^+} + \beta \frac{-\nabla_\theta f_\theta^-}{1-f_\theta^-} \Bigr)\\
        & = - \mathbb{E}_\mathcal{D} \Bigl( \frac{\nabla_\theta f_\theta}{u} + \beta \frac{-\nabla_\theta f_\theta}{1-u-\epsilon} \Bigr)\\
        & = - \beta \mathbb{E}_\mathcal{D} \Bigl[ \frac{1-(1+\beta)u-\epsilon}{u(1-u-\epsilon)} \nabla_\theta f_\theta \Bigr]\\
    \end{split}
\end{equation}

\begin{equation}
    \begin{split}
        \nabla_\theta \mathcal{L}_{ExMATE} & = - \mathbb{E}_\mathcal{D} \Bigl( \frac{\nabla_\theta f_\theta^+}{f_\theta^+} -\beta\nabla_\theta f_\theta^- \Bigr)\\
        & =  - \mathbb{E}_\mathcal{D} \Bigl( \frac{\nabla_\theta f_\theta}{u} -\beta\nabla_\theta f_\theta \Bigr)\\
        & =  - \mathbb{E}_\mathcal{D} \Bigl( \frac{1}{u} - \beta \Bigr) \nabla_\theta f_\theta
    \end{split}
\end{equation}
    \section{Experiment Details}
We use CausalDialogue dataset (\url{https://github.com/Pascalson/CausalDialogue/tree/main/data}) under GNU Free Document License and Anthropic HH-RLHF dataset (\url{https://huggingface.co/datasets/Anthropic/hh-rlhf}) under MIT Lisense.

We use single NVIDIA RTX A6000 for training each model on CausalDialogue and comsume around 20G GPU memory and 5-10 hours.
We use four NVIDIA A100 to fine-tune each Pythia2.8B model on Anthropic HH-RLHF and consume around 39G memory per GPU and 2-4 hours; We use eight NVIDIA A100 to fine-tune each Pythia6.9B model and consume around 35G memory per GPU and 3-10 hours.
The time consuming range is due to that DPO takes around twice memory and operations than other methods and may need to change its used batch size.
DPO in the end takes around 1.5-3 times of training time per model.

The error bars we shown in figures are 95\% confidence interval.

The GPT4 judgement experiment is conducted using the gpt-4-turbo-2024-04-09 model version. We give the model a conversation between {\it Human} and {\it Assistant} and two options {\it Response1} and {\it Response2}. We then instruct the model ``You are the human in the conversation. Tell me which response you prefer.''
We further verify the GPT4 judgement by asking participating humans the same task.

\section{Experiments: Sensitivity to hyper-parameters}

\begin{figure}[h]
    \centering
    \includegraphics[width=.98\linewidth]{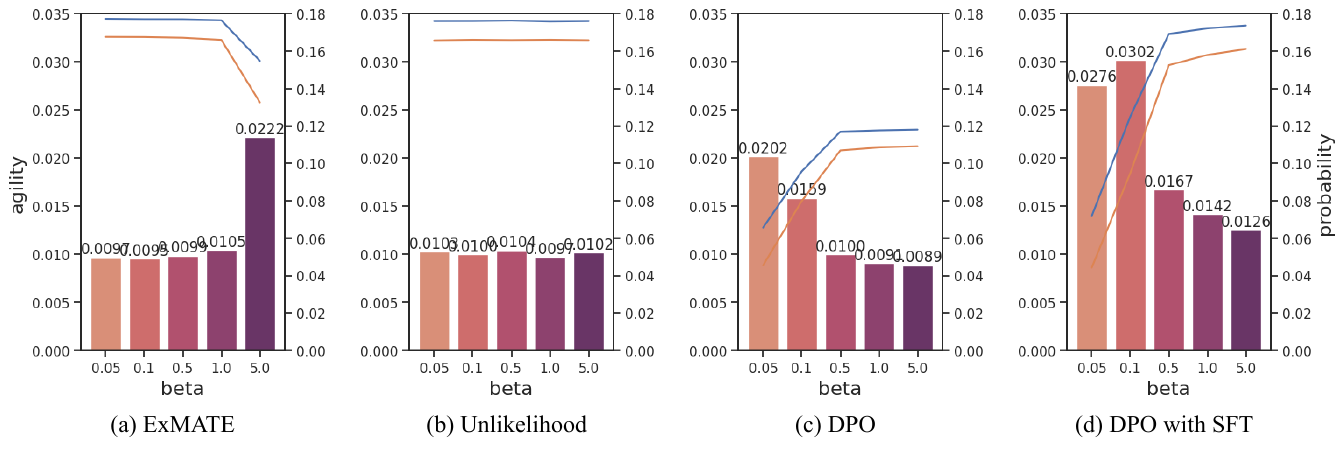}
    \caption{(a)(b)(c) Positive examples probability and ATE agility using ExMATE, Unlikelihood, and DPO to finetune Pythia-2.8B model without supervised fine-tuning. (d) Results of SFT+DPO.}
    \label{fig:beta-sweeps}
\end{figure}

We further test how the coefficient $\beta$ in all RGPB methods impact their optimization results, as we discussed in Section~\ref{sec:grad-analysis} that $\beta$ can impact the gradients much.
The results are plotted in Figure~\ref{fig:beta-sweeps}.
As we raise $\beta$, the probabilities of ExMATE decay while the agility increases.
In contrast, as $\beta$ rises, the agility of DPO (regardless of the SFT stage) decays and the probabilities are less deteriorated.
This is as the gradient analysis that higher $\beta$ switch ExMATE from having mostly positive gradients to have some negative gradient, therefore decreasing the probabilities.
Meanwhile, higher $\beta$ makes DPO to have smaller negative gradient, thus reducing the rate of decreasing the probabilities.
    
\end{document}